\documentclass[runningheads]{llncs}
\usepackage{graphicx}
\usepackage{tikz}
\usepackage{comment}
\usepackage{amsmath,amssymb} % define this before the line numbering.
\usepackage{color}
\usepackage{booktabs}
\usepackage{multirow}
\usepackage{amssymb}
\usepackage[colorlinks,
            citecolor=green,        
            ]{hyperref}
\usepackage{appendix}
\usepackage{algorithm}
\usepackage{algorithmic}
\usepackage{subcaption}
\usepackage{wrapfig}

\usepackage[accsupp]{axessibility}  % Improves PDF readability for those with disabilities.

\newcommand{\shanshan}[1]{\textcolor[rgb]{1,0,0}{\textbf{Shanshan}: #1}}
\newcommand{\ligang}[1]{\textcolor{black}{#1}}

\begin{document}
% \linenumbers
\pagestyle{headings}
\mainmatter
\def\ECCVSubNumber{2783}  % Insert your submission number here

\title{PseCo: Pseudo Labeling and Consistency Training for Semi-Supervised Object Detection} 

% \begin{comment}
\titlerunning{ECCV-22 submission ID \ECCVSubNumber} 
\authorrunning{ECCV-22 submission ID \ECCVSubNumber} 
\author{Anonymous ECCV submission}
\institute{Paper ID \ECCVSubNumber}
% \end{comment}
%******************

\titlerunning{PseCo for Semi-Supervised Object Detection}
\newcommand*\samethanks[1][\value{footnote}]{\footnotemark[#1]}
\author{Gang Li\inst{1,2} \and
Xiang Li\inst{1}\thanks{Corresponding author.} \and
Yujie Wang\inst{2} \and 
Yichao Wu\inst{2} \and 
Ding Liang\inst{2} \and \\
Shanshan Zhang\inst{1}\samethanks}
\authorrunning{Li et al.}
\institute{Nanjing University of Science and Technology \and
SenseTime Research \\
\email{\{gang.li, xiang.li.implus, shanshan.zhang\}@njust.edu.cn} \\
\email{\{wangyujie,wuyichao,liangding\}@sensetime.com}
}

%******************
\maketitle

\begin{abstract}
In this paper, we delve into two key techniques in Semi-Supervised Object Detection (SSOD), namely pseudo labeling and consistency training. We observe that these two techniques currently neglect some important properties of object detection, hindering efficient learning on unlabeled data.
Specifically, for pseudo labeling, existing works only focus on the classification score yet fail to guarantee the localization precision of pseudo boxes; 
%As a result, noisy pseudo boxes can mislead the label assign and regression task. 
For consistency training, the widely adopted random-resize training only considers the label-level consistency but misses the feature-level one, which also plays an important role in ensuring the scale invariance.
To address the problems incurred by noisy pseudo boxes, we design Noisy Pseudo box Learning (NPL) that includes Prediction-guided Label Assignment (PLA) and Positive-proposal Consistency Voting (PCV). 
PLA relies on model predictions to assign labels and makes it robust to even coarse pseudo boxes; while PCV leverages the regression consistency of positive proposals to reflect the localization quality of pseudo boxes.
%contains two approaches: a prediction-guided label assign strategy, which assigns labels based on model predictions and makes it robust to coarse pseudo boxes;
%and a consistency voting method, which leverages the regression consistency of positive proposals to represent the localization quality of pseudo boxes.       
%In consistency training, we realize feature consistency between different input scales via aligning their features, which can by easily implemented by shifting pyramid feature position.
Furthermore, in consistency training, we propose Multi-view Scale-invariant Learning (MSL) that includes mechanisms of both label- and feature-level consistency, where feature consistency is achieved by aligning shifted feature pyramids between two images with identical content but varied scales. 
On COCO benchmark, our method, termed PSEudo labeling and COnsistency training (PseCo), outperforms the SOTA (Soft Teacher) by 2.0, 1.8, 2.0 points under 1\%, 5\%, and 10\% labelling ratios, respectively.  
It also significantly improves the learning efficiency for SSOD, e.g., PseCo halves the training time of the SOTA approach but achieves even better performance. Code is available at \href{https://github.com/ligang-cs/PseCo}{\color{blue}{https://github.com/ligang-cs/PseCo}}.
% \shanshan{the top approach or several SOTA approaches?}

\end{abstract}

\section{Introduction}

% 半监督是一个很有前景的研究方向
With the rapid development of deep learning, many computer vision tasks achieve significant improvements, such as image classification~\cite{imagenet}, object detection~\cite{swin,mixmatch,RM}, etc. Behind these advances, plenty of annotated data plays an important role~\cite{Soco}. However, labeling accurate annotations for large-scale data is usually time-consuming and expensive, especially for object detection, which requires annotating precise bounding boxes for each instance, besides category labels. Therefore, employing easily accessible unlabeled data to facilitate the model training with limited annotated data is a promising direction, named Semi-Supervised Learning, where labeled data and unlabeled data are combined together as training examples.  

% 阐述一下现在的半监督框架是怎么做的；但是这种框架直接用于检测，并不是最优解
Semi-Supervised for Image Classification (SSIC) has been widely investigated in previous literature, and the learning paradigm on unlabeled data can be roughly divided into two categories: pseudo labeling~\cite{pseudo-label,pseudo-label2} and consistency training~\cite{UDA,MeanTeacher}, each of which receives much attention. Recently, some works (e.g., FixMatch~\cite{Fixmatch}, FlexMatch~\cite{Flexmatch}) attempt to combine these two techniques into one framework and achieve state-of-the-art performance. 
In Semi-Supervised Object Detection (SSOD), some works borrow the key techniques (e.g. pseudo labeling, consistency training) from SSIC, and directly apply them to SSOD. 
Although these works~\cite{instantTeaching,softTeacher} obtain gains from unlabeled data, they neglect some important properties of object detection, resulting in sub-optimal results.     
On the one hand, compared with image classification, pseudo labels of object detection are more complicated, containing both category and location information. On the other hand, object detection is required to capture stronger scale-invariant ability than image classification, as it needs to carefully deal with the targets with rich scales.
In this work, we present a SSOD framework, termed PSEudo labeling and COnsistency training (PseCo), to integrate object detection properties into SSOD, making pseudo labeling and consistency training work better for object detection tasks.             

\begin{figure}[t]
    \centering
    \includegraphics[width=0.95\textwidth]{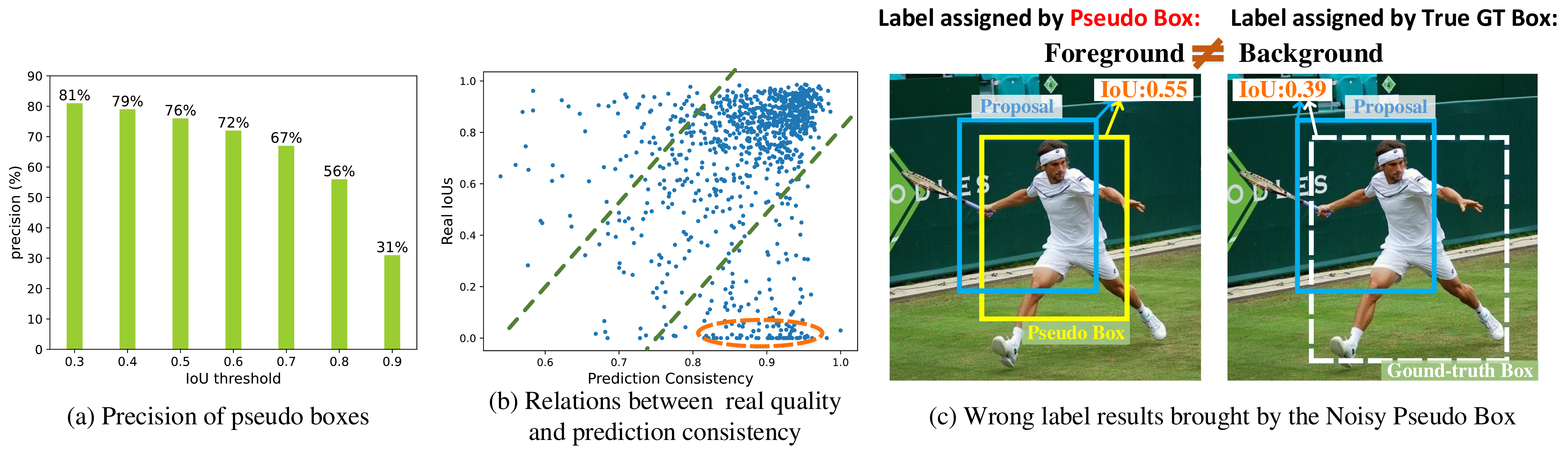}
    \vspace{-7pt}
    \caption{(a) The precision of pseudo boxes under various IoU thresholds. (b) The scatter diagram of the relation between the prediction consistency and their true localization quality. Some dots falling in the orange ellipse are caused by annotation errors. We show some examples in Fig.~\ref{fig:qualitative}. (c) One specific example to demonstrate that noisy pseudo boxes will mislead label assignment.}
    \label{fig:motivation}
\end{figure}

In pseudo labeling, the model produces one-hot pseudo labels on unlabeled data by itself, and only pseudo labels whose scores are above the predefined score threshold are retained.  
As for object detection, the pseudo label consists of both category labels and bounding boxes. Although category labels can be guaranteed to be accurate via setting a high score threshold, the localization quality of pseudo box fails to be measured and guaranteed. 
It has been validated in previous works that the classification score is not strongly correlated with the precision of box localization~\cite{gfl,varifocalnet,IoUNet,softTeacher}. 
In Fig.~\ref{fig:motivation}(a), we compute the precision of pseudo boxes under various Intersection-over-Union (IoU) thresholds, via comparing produced pseudo boxes with ground-truths.  
Under loose criterion (IoU=0.3), precision can reach 81\%, but it will drop to 31\% when we lift the IoU threshold to 0.9. This dramatic precision gap indicates coarse pseudo boxes whose IoUs belong to [0.3,0.9] account for 50\%. 
% \shanshan{how can we see it is half?}
If these noisy pseudo boxes are used as targets to train the detector, it must hinder the optimization, resulting in slow convergence and inefficient learning on unlabeled data.
Furthermore, we analyze the negative effects brought by noisy pseudo boxes on classification and regression tasks as follows, respectively. 

For the classification task, noisy pseudo boxes will mislead the label assignment, where labels are assigned based on IoUs between proposals and gt boxes (pseudo boxes in our case). 
As shown in Fig.~\ref{fig:motivation}(c), a background proposal is taken as foreground due to a large IoU value with a poorly localized pseudo box.
% This label assign strategy relies on the assumption that ground truth boxes are precise, however, this assumption does not hold for pseudo boxes obviously. 
As a result, the IoU-based label assignment will fail on unlabeled data and confuse decision boundaries between foreground and background.
% One specific example is shown in Fig.~\ref{fig:motivation}(c).
To address this issue, we design a prediction-guided label assignment strategy for unlabeled data, which assigns labels based on predictions of the teacher, instead of IoUs with pseudo boxes as before, making it robust for poorly localized pseudo boxes. 
% \implusrewrite{add words to explain robust.}

For the regression task, it is necessary to measure the localization quality of pseudo boxes. 
We propose a simple yet effective method to achieve this, named Positive-proposal Consistency Voting. 
We empirically find that regression consistency from positive proposals is capable of reflecting the localization quality of corresponding pseudo boxes. In Fig.~\ref{fig:motivation}(b), we visualize the relations between predicted consistency and their true IoUs, where their positive correlations can be found. Therefore, it is reasonable to employ the estimated localization quality (i.e., regression consistency from positive proposals) to re-weight the regression losses, making precise pseudo boxes contribute more to regression supervisions. 
%Then, estimated localization quality will be employed to re-weight the regression losses, making precise pseudo boxes contribute more to regression supervisions.   

Apart from pseudo labeling, we also analyze the consistency training for SSOD. Consistency training enforces the model to generate similar predictions when fed with perturbed versions of the same image, where perturbations can be implemented by injecting various data augmentations. Through consistency training, models can be invariant to different input transformations. 
Current SSOD methods~\cite{instantTeaching,softTeacher,unbiasedTeacher} only apply off-the-shelf, general data augmentations, most of which are borrowed from image classification.
However, different from classification, object detection is an instance-based task, where object scales usually vary in a large range, and detectors are expected to handle all scale ranges. Therefore, learning strong scale-invariant ability via consistency training is important.
In scale consistency, it should be allowed for the model to predict the same boxes for input images with identical contents but varied scales.
% The goal of scale consistency is to allow the model to predict the same boxes for input images with identical contents but varied scales.
To ensure label consistency, random-resizing is a common augmentation, which resizes input images and gt boxes according to a randomly generated resize ratio. 
% \implusrewrite{which is } % too many "which"
Besides label consistency, feature consistency also plays an important role in scale-invariant learning, but it is neglected in previous works.
% Exactly, through aligning features for input images with identical content but varied scales, these images can predict the same boxes, realizing feature consistency. 
Thanks to the pyramid structure of popular backbone networks, feature alignment can be easily implemented by shifting feature pyramid levels according to the scale changes.
% Thanks to the pyramid structure of detectors, feature alignment can be easily implemented by shifting feature pyramid position, when input scales differ in even number times (e.g., 2x, 4x).    
Motivated by this, we introduce a brand new data augmentation technique, named Multi-view Scale-invariant Learning (MSL), to learn label-level and feature-level consistency simultaneously in a simple framework.
%at the same time. 
        
In summary, we delve into two key techniques of semi-supervised learning (e.g., pseudo labeling and consistency training) for SSOD, and integrate object detection properties into them. 
On COCO benchmarks, our PseCo outperforms the state-of-the-art methods by a large margin, for example, under 10\% labelling ratio, it can improve a 26.9\% mAP baseline to 36.1\% mAP, surpassing previous methods by at least 2.0\%. When labeled data is abundant, i.e., we use full COCO training set as labeled data and extra 123K unlabeled2017 as unlabeled data, our PseCo improves the 41.0\% mAP baseline by +5.1\%, reaching 46.1\% mAP, establishing a new state of the art. Moreover, PseCo also significantly boosts the convergence speed, e.g. PseCo halves the training time of the SOTA (Soft Teacher~\cite{softTeacher}), but achieves even better performance.
% \implusrewrite{say something about the efficiency.}

\section{Related Works}

\noindent\textbf{Semi-supervised learning in image classification.}
Semi-supervised learning can be categorized into two groups: pseudo labeling (also called self-training) and consistency training, and previous methods design learning paradigms based on one of them. Pseudo labeling~\cite{pseudo-label,pseudo-label2,pseudo_label_3,NoisyStudent} iteratively adds unlabeled data into the training procedure with pseudo labels annotated by an initially trained network. Here, only model predictions with high confidence will be transformed into the one-hot format and become pseudo labels. Noisy Student Training~\cite{NoisyStudent} injects noise into unlabeled data training, which equips the model with stronger generalization through training on the combination of labeled and unlabeled data. 
On the other hand, consistency training~\cite{MeanTeacher,UDA,mixmatch} relies on the assumption that the model should be invariant to small changes on input images or model hidden states. It enforces the model to make similar predictions on the perturbed versions of the same image, and perturbations can be implemented by injecting noise into images and hidden states. 
UDA~\cite{UDA} validates the advanced data augmentations play a crucial role in consistency training, and observes the strong augmentations found in supervised-learning can also lead to obvious improvements in semi-supervised learning.

Recently, some works~\cite{Fixmatch,Flexmatch} attempt to combine pseudo labeling and consistency training, achieving state-of-the-art performance. 
FixMatch~\cite{Fixmatch} firstly applies the weak and strong augmentations to the same input image, respectively, to generate two versions, then uses the weakly-augmented version to generate hard pseudo labels. The model is trained on strongly-augmented versions to align predictions with pseudo labels.
Based on FixMatch, FlexMatch~\cite{Flexmatch} proposes to adjust score thresholds for different classes during the generation of pseudo labels, based on curriculum learning.
It has been widely validated that pseudo labeling and consistency training are two powerful techniques in semi-supervised image classification, hence in this work, we attempt to integrate object detection properties into them and make them work better for semi-supervised object detection. 

\noindent\textbf{Semi-supervised learning in object detection.} STAC~\cite{STAC} is the first attempt to apply pseudo labeling and consistency training based on the strong data augmentations to semi-supervised object detection, however, it adopts two stages of training as Noisy Student Training~\cite{NoisyStudent}, which prevents the pseudo labels from updating along with model training and limits the performance.
After STAC, ~\cite{softTeacher,instantTeaching,humbleTeacher,ISMT,unbiasedTeacher} borrow the idea of Exponential Moving Average (EMA) from Mean Teacher~\cite{MeanTeacher}, and update the teacher model after each training iteration to generate instant pseudo labels, realizing the end-to-end framework. To pursue high quality of pseudo labels and overcome confirmation bias, Instant-Teaching~\cite{instantTeaching} and ISMT~\cite{ISMT} introduce model ensemble to aggregate predictions from multiple teacher models which are initialized differently; similarly, Humble Teacher~\cite{humbleTeacher} ensembles the teacher model predictions by taking both the image and its horizontally flipped version as input. Although these ensemble methods can promote the quality of pseudo labels, they also introduce considerable computation overhead. 
Unbiased Teacher~\cite{unbiasedTeacher} replaces traditional Cross-entropy loss with Focal loss~\cite{focal} to alleviate the class-imbalanced pseudo-labeling issue, which shows strong performance when labeled data is scarce. 
Soft Teacher~\cite{softTeacher} uses teacher classification scores as classification loss weights, to suppress negative effects from underlying objects missed by pseudo labels.
Different from previous methods, our work elaborately analyzes whether the pseudo labeling and consistency training can be directly applied to SSOD, but gets a negative answer. 
To integrate object detection properties into these two techniques, we introduce Noisy Pseudo box Learning and Multi-view Scale-invariant Learning, obtaining much better performance and faster convergence speed.   

\section{Method}

\begin{figure}[t]
    \centering
    \includegraphics[width=0.9\textwidth]{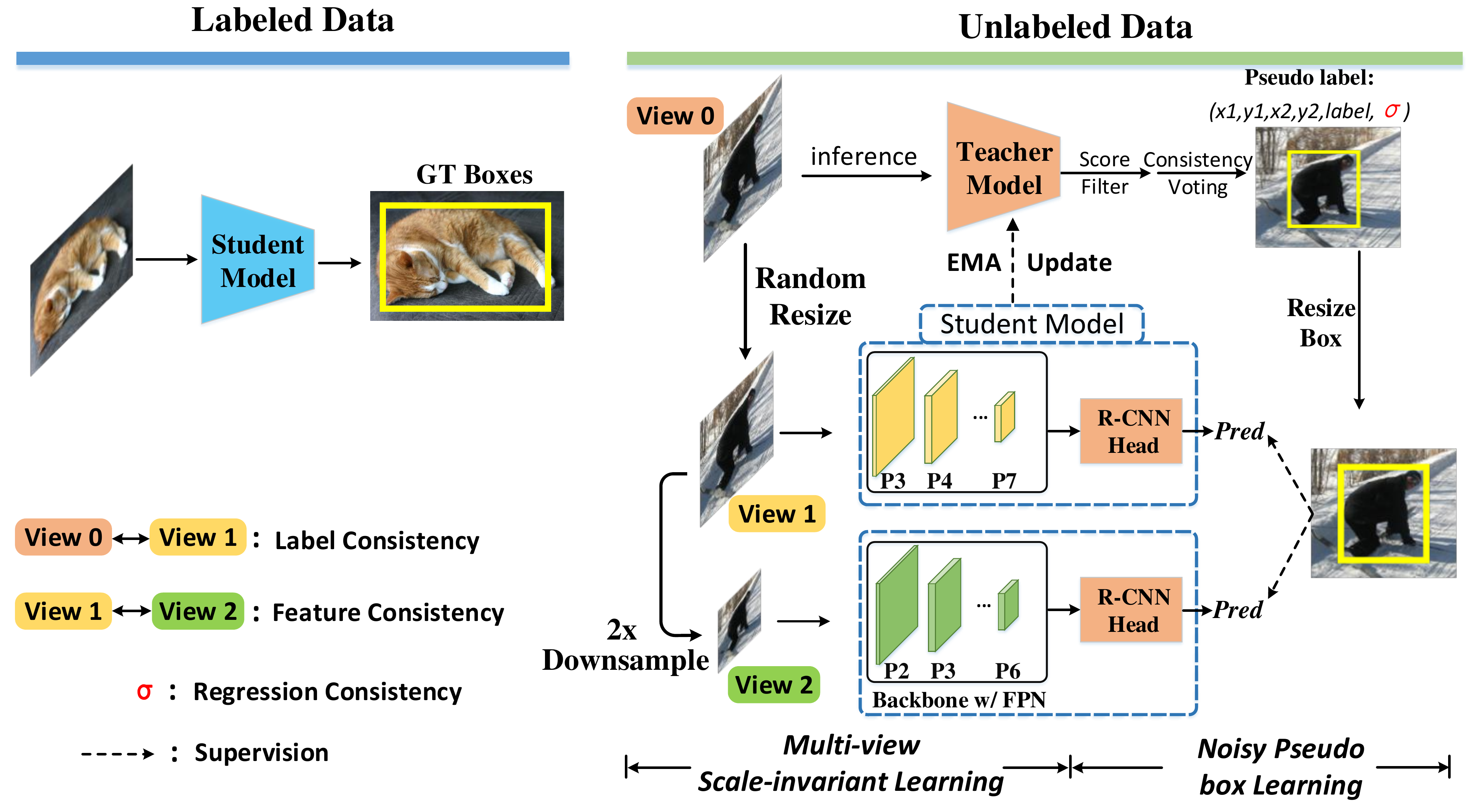}
    \vspace{-10pt}
    \caption{The framework of our PseCo. 
    % Labeled data and unlabeled data form the training batch. 
    Each training batch consists of both labeled and unlabeled images.
    On the unlabeled images, the student model trains on view $V_{1}$ and $V_{2}$ at the same time, taking the same pseudo boxes as supervisions. View $V_{0}$ refers to input images for the teacher model.
    }
    \label{fig:framework}
\end{figure}

We show the framework of our PseCo in Fig.~\ref{fig:framework}.
On the unlabeled data, PseCo consists of Noisy Pseudo box Learning (NPL) and Multi-view Scale-invariant Learning (MSL). In the following parts, we will introduce the basic framework, the proposed NPL and MSL, respectively.  

\subsection{The basic framework}

At first, we directly apply standard pseudo labeling and consistency training to SSOD, building our basic framework.
Following previous works~\cite{softTeacher,unbiasedTeacher,instantTeaching}, we also adopt Teacher-student training scheme, where the teacher model is built from the student model at every training iteration via Exponential Moving Average (EMA).
We randomly sample labeled data and unlabeled data based on a sample ratio to form the training batch. On the labeled data, the student model is trained in a regular manner, supervised by the ground-truth boxes:
\begin{equation}
    \mathcal{L}^{l} = \mathcal{L}^{l}_{cls} + \mathcal{L}^{l}_{reg}.
\end{equation}
On the unlabeled data, we firstly apply weak data augmentations (e.g. horizontal flip, random resizing) to input images, and then feed them to the teacher model for pseudo label generation. 
Considering the detection boxes tend to be dense even after NMS, we set a score threshold $\tau$ and only retain boxes with scores above $\tau$ as pseudo labels.
After that, strong augmentations (e.g. cutout, rotation, brightness jitter)\footnote{We adopt the same data augmentations as Soft Teacher~\cite{softTeacher}, please refer to~\cite{softTeacher} for more augmentation details.} will be performed on the input image to generate the training example for student model. Since high classification scores do not lead to precise localization, we abandon bounding box regression on unlabeled data, as done in~\cite{unbiasedTeacher}.
Actually, applying the box regression loss on unlabeled data will cause unstable training in our experiments.

Foreground-background imbalance~\cite{ghm,focal} is an intrinsic issue in object detection, and it gets worse under the semi-supervised setting. A high score threshold $\tau$ is usually adopted to guarantee the precision of pseudo labels, but it also results in scarcity of pseudo labels, aggravating the imbalance of foreground/background.
Moreover, there also exists foreground-foreground imbalance, exactly, training examples from some specific categories can be limited when labeled data is scarce, which makes the model prone to predict the dominant classes, causing biased prediction. To alleviate these imbalance issues, we follow the practice of Unbiased Teacher~\cite{unbiasedTeacher}, and replace the standard cross-entropy loss with focal loss~\cite{focal}:
\begin{equation}
    \mathcal{L}^{u}_{cls} = -\alpha_{t}(1-p_{t})^{\gamma}log(p_{t}), \ 
    p_{t} = \begin{cases}
    p, &if\ y = 1, \\
    1 - p, &otherwise,
    \end{cases}
\end{equation}
where parameters $\alpha_{t}$ and $\gamma$ adopt default settings in original %Focal-Loss
focal loss
paper~\cite{focal}. The overall loss function is formulated as:
\begin{equation}
    \mathcal{L} = \mathcal{L}^{l} + \beta\mathcal{L}^{u},
\end{equation}
where $\beta$ is used to control the contribution of unlabeled data. 
In theory, our proposed method is independent of the detection framework and can be applied on both one-stage and two-stage detectors. However, considering all previous methods are based on Faster R-CNN~\cite{faster} detection framework, for a fair comparison with them, we also adopt Faster R-CNN as the default detection framework.
% in this work. 
% \shanshan{Please note our proposed method is independent to the detection framework and can be applied on top of any arbitrary one.} --addressed

\subsection{Noisy Pseudo Box Learning}
% \shanshan{Tolerant}

In SSOD, pseudo labels contain both category and location. Since the score of pseudo labels can only indicate the confidence of pseudo box categories, the localization quality of pseudo boxes is not guaranteed. Imprecise pseudo boxes will mislead the label assignment and regression task, making learning on unlabeled data inefficient.
Motivated by this, we introduce Prediction-guided Label Assignment and Positive-proposal Consistency Voting to reduce negative effects on the label assignment and regression task, respectively.

\noindent\textbf{Prediction-guided Label Assignment.} The standard label assignment strategy in Faster R-CNN~\cite{faster} only takes the IoUs between proposals and gt boxes (pseudo boxes in our case) into consideration and assigns foreground to those proposals, whose IoUs are above a pre-defined threshold $t$ (0.5 as default). This strategy relies on the assumption that gt boxes are precise, however, this assumption does not hold for unlabeled data obviously. As a result, some low-quality proposals will be mistakenly assigned as positive, confusing the classification boundaries between foreground and background. One specific example is shown in Fig.~\ref{fig:motivation}(c), where a proposal with the true IoU as 0.39 is mistakenly assigned as positive.  

To address this problem, we propose Prediction-guided Label Assignment (PLA), which takes teacher predictions as auxiliary information and reduces dependency on IoUs. In Teacher-student training scheme, 
not only can the detection results (after NMS) of teacher perform as pseudo labels, but also teacher's dense predictions (before NMS) are able to provide guidance for student model training. 
We share the proposals generated by the teacher RPN with the student, so that teacher predictions on these proposals can be easily transferred to student. 
\ligang{To measure the proposal quality ($q$) comprehensively, the classification confidence and localization precision of teacher predictions are jointly employed, concretely, $q = s^{\alpha} \times u^{1 - \alpha}$, where $s$ and $u$ denote a foreground score and an IoU value between the regressed box and the ground truth, respectively. $\alpha$ controls the contribution of $s$ and $u$ in the overall quality.} 
% For proposals in teacher network, we leverage their classification scores (denoted as $p$) to represent classification confidence, and IoUs (denoted as $q$) between their regressed boxes and pseudo boxes to represent regression quality.
\ligang{On unlabeled data, we first construct a candidate bag for each ground truth $g$ by the traditional IoU-based strategy, where the IoU threshold $t$ is set to a relatively low value, e.g., 0.4 as default, to contain more proposals.
Within each candidate bag, the proposals are firstly sorted by their quality $q$, then top-$\mathcal{N}$ proposals are adopted as positive samples and the rest are negatives. The number $\mathcal{N}$ is decided by \textit{the dynamic k estimation} strategy proposed in OTA~\cite{ota}, specifically, the IoU values over the candidate bag is summed up to represent the number of positive samples. 
}
% On unlabeled data, we first lower the IoU threshold $t$ from 0.5 to 0.4, in order to construct bigger candidate bags containing more candidate proposals for each pseudo box (indexed by $j$). Then among these candidate proposals, we select top $k$ proposals as positive, according to proposal quality, which is defined as $quality = \sqrt{pq}$ that simultaneously considers both classification and localization aspects. Here, the number of positive proposals $k$ is determined by Dynamic k Estimation used in OTA~\cite{ota}, which can be formulated as $k^{j} = \lfloor \sum_{i=1}^{N}q_{i}^{j} \rfloor$, where $i$ refers to the proposal index in the candidate bag belonging to pseudo box $j$.  
The proposed PLA gets rid of strong dependencies on IoUs and alleviates negative effects from poorly localized pseudo boxes, leading to clearer classification boundaries.  
%From another perspective, 
Furthermore, our label assign strategy integrates more teacher knowledge into student model training, realizing better knowledge distillation.

\noindent\textbf{Positive-proposal Consistency Voting.} Considering the classification score fails to indicate localization quality, we introduce a simple yet effective method to measure the localization quality, named Positive-proposal Consistency Voting (PCV).
Assigning multiple proposals to each gt box (or pseudo box) is a common practice in CNN-based detectors~\cite{faster,gfl,varifocalnet}, and we observe that the consistency of regression results from these proposals is capable of reflecting the localization quality of the corresponding pseudo box. Regression consistency $\sigma^{j}$ for pseudo box (indexed by $j$) is formulated as:
\begin{equation}
    \sigma^{j} = \frac{\sum_{i=1}^{N}u_{i}^{j}}{N},
    \label{eq:consistency}
\end{equation}
where $u$ denotes an IoU value between the predicted box and the pseudo box, as defined above; $N$ denotes the number of positive proposals, assigned to the pseudo box $j$. After obtaining $\sigma^{j}$, we employ it as the instance-wise regression loss weight:
\begin{equation}
    \mathcal{L}^{u}_{reg} = \frac{1}{MN}\sum_{j=1}^M\sigma^{j}\sum_{i=1}^N|reg_{i}^{j}-\hat{reg}_{i}^{j}|,
\end{equation}
where $reg$ and $\hat{reg}$ refer to the regression output and ground-truth, respectively. In Fig.~\ref{fig:motivation}(b), we depict the scatter diagram of the relation between prediction consistency $\sigma$ of pseudo boxes and their true IoUs. It is obvious that $\sigma$ is positively correlated with true IoUs. Note that, some dots falling in the orange ellipse are mainly caused by annotation errors. We visualize some examples in Fig.~\ref{fig:qualitative}, 
where the pseudo boxes accurately detect some objects, which are missed by the ground truths.      

\begin{figure}[t]
    \centering
    \includegraphics[width=0.9\textwidth]{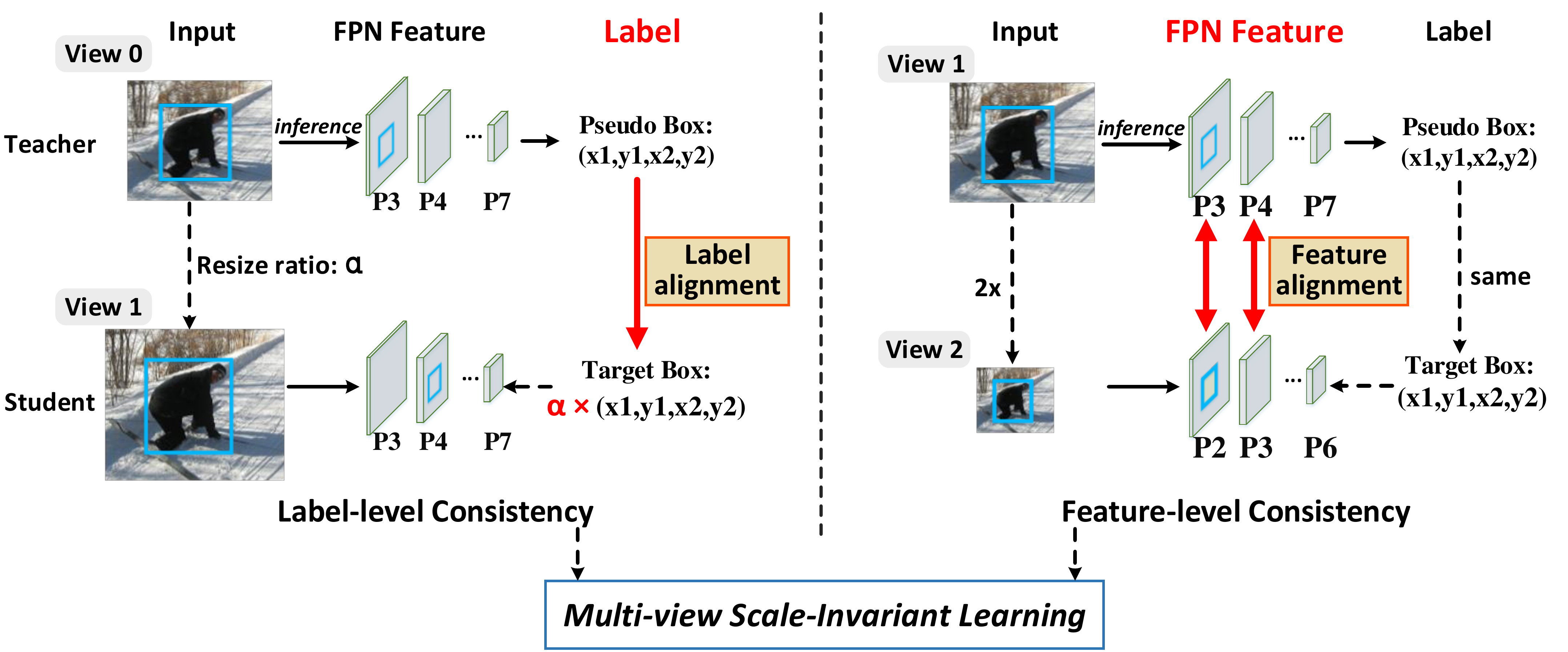}
    \vspace{-7pt}
    \caption{Comparisons between label-level consistency learning and feature-level consistency learning. 
    % To realize consistency, in label consistency,
    For label consistency,
    labels are aligned according to the resize ratio $\alpha$; for feature consistency, features are aligned by shifting 
    the feature pyramid level.
    % pyramid feature position.
    }
    \label{fig:MSL}
\end{figure}

\subsection{Multi-view Scale-invariant Learning}
% \shanshan{or Invariant?}
% \shanshan{The two paragraphs need to be re-organized. Let's talk about the motivation in the 1st one and our method in the 2nd one. Currently, the 2nd one is a mixture.} -- addressed
Different from image classification, in object detection, object scales vary in a large range and detectors hardly show comparable performance on all scales. Therefore, learning scale-invariant representations from unlabeled data is considerably important for SSOD. 
In consistency training, strong data augmentations play a crucial role~\cite{UDA,NoisyStudent} in achieving competitive performance. Through injecting the perturbations into the input images, data augmentations equip the model with robustness to various transformations. From the perspective of scale invariance, we regard the common data augmentation strategy (e.g. random-resizing) as label-level consistency since it resizes the label according to the scale changes of input images. 
Unfortunately, existing works only involve the widely adopted label-level consistency but fail to consider the feature-level one. 
Since detection network usually has designs of rich feature pyramids, feature-level consistency is easy to implement across paired inputs~\cite{low-resKD} and should be considered seriously. 
In this paper, we propose Multi-view Scale-invariant Learning (MSL) that combines both label- and feature-level consistency into a simple framework, where feature-level consistency is realized by aligning shifted pyramid features between two images with identical content but different scales.
% \implusrewrite{briefly introduce the method, refer to abstract}.
%In this work, we delve deep into scale-invariant learning, 
%and observe scale-invariant learning actually contains two components, 
%namely label-level consistency and feature-level consistency, but the latter is neglected by previous works. 
%Label-level consistency (noted as label consistency for short) is usually implemented by random-resize, where gt boxes are resized according to scale changes of input images, in order to maintain the consistency between input and label.
%On the other hand, feature consistency is seldom investigated in previous works~\cite{low-resKD}. Specifically, taking two images with identical content but varying scales as inputs, the network will produce different feature sizes for the same region. 
%As a result, different boxes should be predicted from this region. 
%Only when we align feature sizes for these images with different scales, the network can be capable of predicting the same boxes for them. Thanks to pyramid structure of detectors, the feature alignment can be easily implemented by shifting pyramid feature position. To combine label- and feature-level consistency into the same framework, we propose a brand new augmentation technique, named Multi-view Scale-invariant Learning (MSL).

To be specific, two views, namely $V_{1}$ and $V_{2}$, are used for student training in MSL. We denote the input image for the teacher model as $V_{0}$.
Views $V_{1}$ and $V_{2}$ are constructed to learn label- and feature-level consistency, respectively. 
Among them, $V_{1}$ is implemented by vanilla random resizing, which rescales the input $V_{0}$ and pseudo boxes according to a resize ratio $\alpha$ randomly sampled from the range $[\alpha_{min}, \alpha_{max}]$ ([0.8, 1.3] as default).  
For feature consistency learning, we firstly downsample $V_{1}$ by even number times (2x as default) to produce $V_{2}$, then combine $V_{1}$ and $V_{2}$ into image pairs. Upsampling is also certainly permitted, but we only perform downsampling here for GPU memory restriction. Because the spatial sizes of adjacent FPN layers always differ by 2x, the P3-P7 layers\footnote{$P_{x}$ refers to the FPN layer whose feature maps are downsampled by $2^{x}$ times.} of $V_{1}$ can align well with P2-P6 layers of $V_{2}$ in the spatial dimension. 
Through feature alignment, the same pseudo boxes can supervise the student model training on both $V_{1}$ and $V_{2}$. Integrating label consistency and feature consistency into consistency learning leads to stronger scale-invariant learning and significantly accelerates model convergence, as we will show later in the experiments. Comparisons between label consistency and feature consistency are shown in Fig.~\ref{fig:MSL}.

Learning scale-invariant representation from unlabeled data is also explored by SoCo~\cite{Soco}. However, we claim there are two intrinsic differences between MSL and SoCo: (1) MSL models scale invariance from both label consistency and image feature consistency, while SoCo only considers object feature consistency. Through aligning \textbf{dense image features} of shifted pyramids between paired images, our MSL can provide more comprehensive and dense supervisory signals than the SoCo, which only performs consistency on \textbf{sparse objects}. 
(2) SoCo implements feature consistency via contrastive learning, which is designed for the pretraining; in contrast, our MSL uses bounding box supervision to implement consistency
learning and can be integrated into the detection task.        

\section{Experiments}
% \shanshan{In this section, ...} 
\subsection{Dataset and Evaluation Protocol}
In this section, we conduct extensive experiments to verify the effectiveness of PseCo on MS COCO benchmark~\cite{coco}. There are two training sets, namely the train2017 set, containing 118k labeled images, and the unlabeled2017 set, containing 123k unlabeled images. The val2017 with 5k images is used as validation set, and we report all experiment results on val2017. The performance is measured by COCO average prevision (denoted as mAP). Following the common practice of SSOD~\cite{STAC}, there are two experimental settings: \textbf{Partially Labeled Data} and \textbf{Fully Labeled Data}, which are described as follows:

\noindent\textbf{Partially Labeled Data.} We randomly sample 1, 2, 5, and 10\% data from train2017 as labeled data, and use the rest as unlabeled. Under each labelling ratio, we report the mean and standard deviation over 5 different data folds. 

\noindent\textbf{Fully Labeled Data.} Under this setting, we take train2017 as the training labeled set and unlabeled2017 as the training unlabeled set. 

\subsection{Implementation Details}
For a fair comparison, we adopt Faster R-CNN~\cite{faster} with FPN~\cite{fpn} as the detection framework, and ResNet-50~\cite{resnet} as the backbone. The confidence threshold $\tau$ is set to 0.5, empirically. We set $\beta$ as 4.0 to control contributions of unlabeled data in the overall losses. The performance is evaluated on the Teacher model. Training details for \textbf{Partially Labeled Data} and \textbf{Fully Labeled Data} are described below:

\noindent\textbf{Partially Labeled Data.} All models are trained for 180k iterations on 8 GPUs. The initial learning rate is set as 0.01 and divided by 10 at 120k and 160k iterations. The training batch in each GPU includes 5 images, where the sample ratio between unlabeled data and labeled data is set to 4:1. 

\noindent\textbf{Fully Labeled Data.} All models are trained for 720k iterations on 8 GPUs. Mini-batch in each GPU is 8 with the sample ratio between unlabeled and labeled data as 1:1. The learning rate is initialized to 0.01 and divided by 10 at 480k and 680k iterations.  

\begin{table}[t]
    \centering
    \caption{Comparisons with the state-of-the-art methods on val2017 set under the \textbf{Partially Labeled Data} and \textbf{Fully Labeled Data} settings.}
    % \vspace{-2mm}
    \renewcommand{\arraystretch}{1.2}
    \resizebox{0.85\textwidth}{!}{
    \begin{tabular}{c|c|c|c|c|c}
    \hline
    \multirow{2}{*}{Method} & \multicolumn{4}{c|}{Partially Labeled Data} & \multirow{2}{*}{Fully Labeled Data} \\ \cline{2-5} 
     & 1\% & 2\% & 5\% & 10\% &  \\ \hline
    Supervised baseline & 12.20$\pm$0.29 & 16.53$\pm$0.12 & 21.17$\pm$0.17 & 26.90$\pm$0.08 & 41.0 \\ \hline
    STAC~\cite{STAC} & 13.97$\pm$0.35 & 18.25$\pm$0.25 & 24.38$\pm$0.12 & 28.64$\pm$0.21 & 39.5 $\xrightarrow{-0.3}$ 39.2 \\
    Humble Teacher~\cite{humbleTeacher} & 16.96$\pm$0.35 & 21.74$\pm$0.24 & 27.70$\pm$0.15 & 31.61$\pm$0.28 & 37.6 $\xrightarrow{+4.8}$ 42.4 \\
    ISMT~\cite{ISMT} & 18.88$\pm$0.74 & 22.43$\pm$0.56 & 26.37$\pm$0.24 & 30.53$\pm$0.52 & 37.8 $\xrightarrow{+1.8}$ 39.6 \\
    Instant-Teaching~\cite{instantTeaching} & 18.05$\pm$0.15 & 22.45$\pm$0.15 & 26.75$\pm$0.05 & 30.40$\pm$0.05 & 37.6 $\xrightarrow{+2.6}$ 40.2 \\
    Unbiased Teacher~\cite{unbiasedTeacher} & 20.75$\pm$0.12 & 24.30$\pm$0.07 & 28.27$\pm$0.11 & 31.50$\pm$0.10 & 40.2 $\xrightarrow{+1.1}$ 41.3 \\
    Soft Teacher~\cite{softTeacher}& 20.46$\pm$0.39 & - & 30.74$\pm$0.08 & 34.04$\pm$0.14 & 40.9 $\xrightarrow{+3.6}$ 44.5 \\ \hline
    PseCo (ours) & \textbf{22.43}$\pm$0.36 & \textbf{27.77}$\pm$0.18 & \textbf{32.50}$\pm$0.08 & \textbf{36.06}$\pm$0.24 &  41.0 $\xrightarrow{\textbf{+5.1}}$ \textbf{46.1}\\ \hline
    \end{tabular}}
    \label{tab:partial_labeled}
\end{table}

\subsection{Comparison with State-of-the-Art Methods}

We compare the proposed PseCo with other state-of-the-art methods on COCO val2017 set. Comparisons under the \textbf{Partially Labeled Data} setting are first conducted, with results reported in Tab.~\ref{tab:partial_labeled}. When labeled data is scarce (i.e., under 1\% and 2\% labelling ratios), our method surpasses the state-of-the-art method, Unbiased Teacher~\cite{unbiasedTeacher}, by 1.7\% and 3.5\%, reaching 22.4 and 27.8 mAP, respectively. 
When more labeled data is accessible, the SOTA method is transferred to Soft Teacher~\cite{softTeacher}. Our method still outperforms it by 1.8\% and 2.0\% under 5\% and 10\% labelling ratios, respectively. Therefore, the proposed method outperforms the SOTAs by a large margin, at least 1.7\%, under all labelling ratios.  
Compared with the supervised baseline, PseCo obtains even better performance with only 2\% labeled data than the baseline with 10\% labeled data, demonstrating the effectiveness of proposed semi-supervised learning techniques.

\begin{figure}[t]
    \centering
    \includegraphics[width=0.9\textwidth]{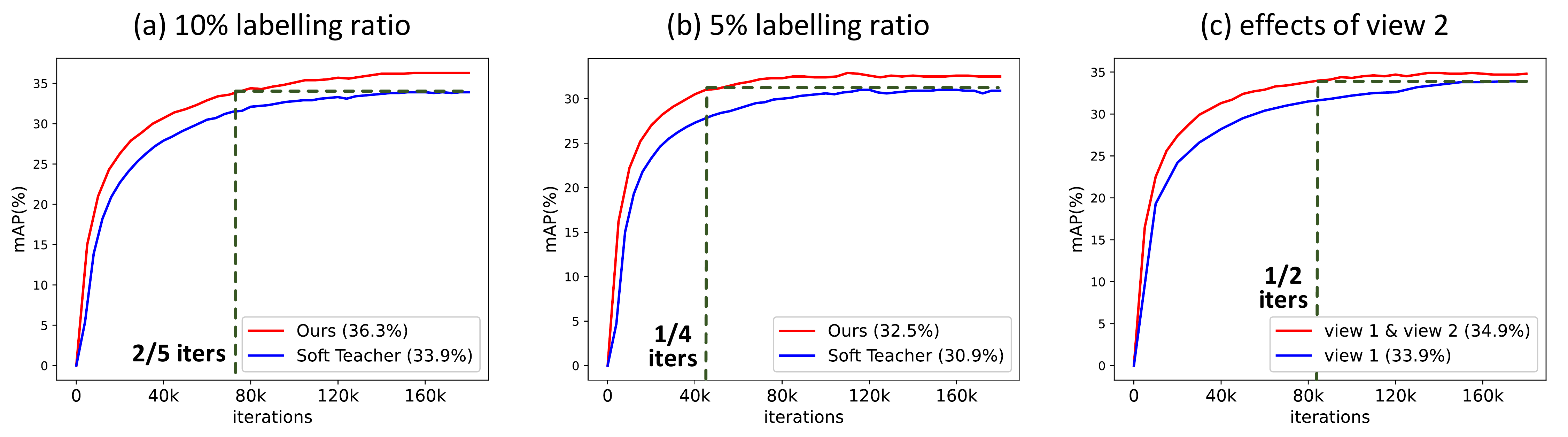}
    \vspace{-3mm}
    \caption{Comparison of model convergence speed.
    In (a) and (b), we compare PseCo against Soft Teacher~\cite{softTeacher}. Here, we reproduce Soft Teacher using their source codes. (c) depicts the comparison between $V_{1}$ and $V_{1}$\&$V_{2}$. In legend, the numbers in brackets refer to mAP. 
    Performance is evaluated on the teacher.
    }
   
    \label{fig:convergence}
\end{figure}

Moreover, we also compare the convergence speed with the previous best method (Soft Teacher~\cite{softTeacher}) in Fig.~\ref{fig:convergence}, where convergence curves are depicted under 10\% and 5\% labelling ratios. It is obvious that our method has a faster convergence speed, specifically, our method uses only 2/5 and 1/4 iterations of Soft Teacher to achieve the same performance under 10\% and 5\% labelling ratios respectively. Although we employ an extra view ($V_{2}$) to learn feature-level consistency, it only increases the training time of each iteration by 25\% (from 0.72 sec/iter to 0.91 sec/iter), due to the low input resolution of $V_{2}$. In summary, we halve the training time of SOTA approach but achieve even better performance, which validates the superior learning efficiency of our method on unlabeled data.  

\begin{figure}[t]
    \centering
    \includegraphics[width=0.9\textwidth]{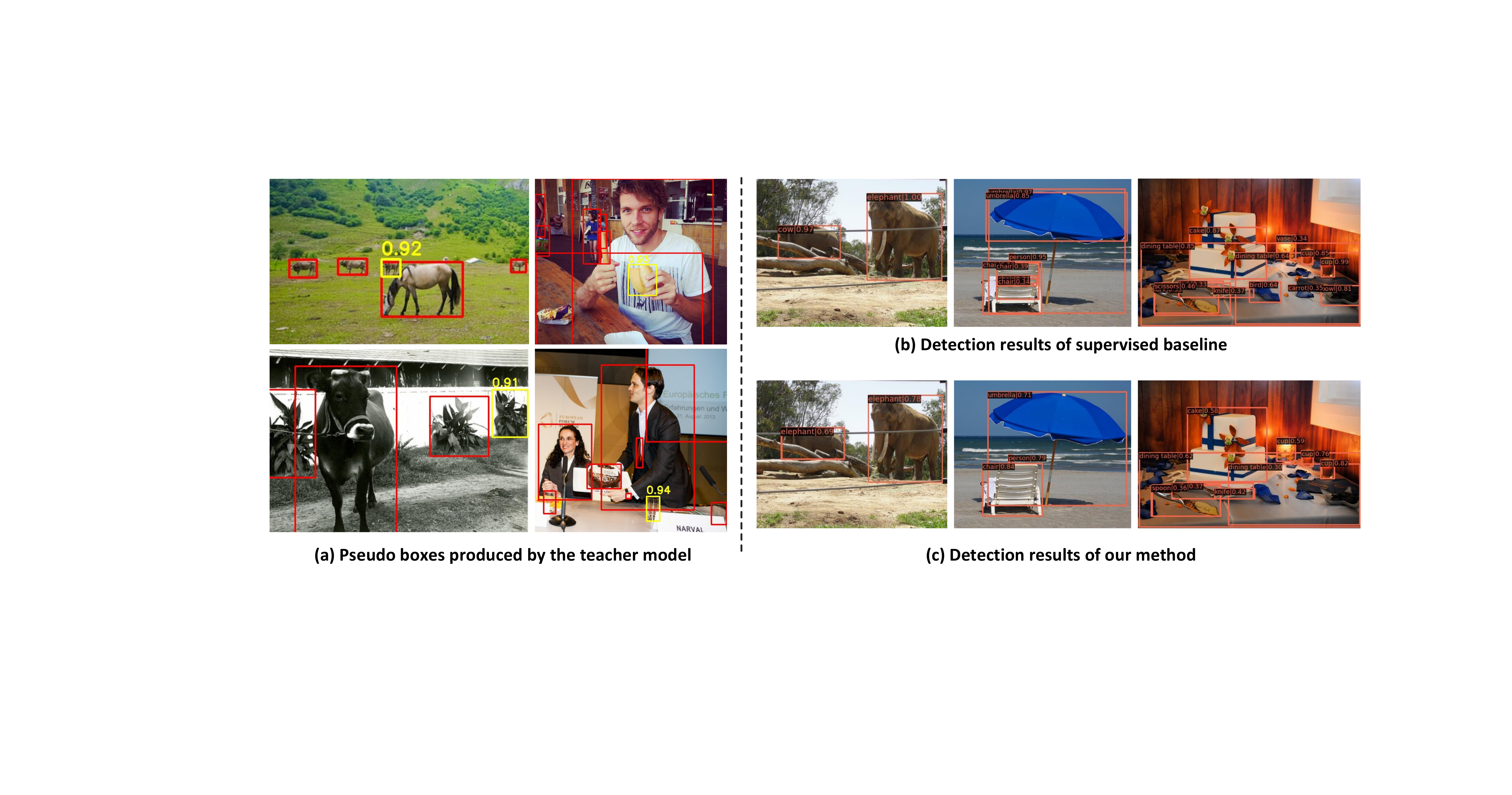}
    \vspace{-3mm}
    \caption{(a) Some pseudo boxes (in yellow) detect objects, missed by ground-truths (in red). Numbers above the pseudo box refer to the predicted consistency $\sigma$. (b)(c) are the results of the supervised baseline and our method.}
    \label{fig:qualitative}
\end{figure}

The experimental results under the \textbf{Fully Labeled Data} setting are reported in Tab.~\ref{tab:partial_labeled}, where both results of comparison methods and their supervised baseline are listed. Following the practice in Soft Teacher~\cite{softTeacher}, we also apply weak augmentations to the labeled data and obtain a strong supervised baseline, 41.0 mAP. Although with a such strong baseline, PseCo still achieves larger improvements (+5.1\%) than others and reaches 46.1 mAP, building a new state of the art. Some qualitative results are shown in Fig.~\ref{fig:qualitative}.    

\begin{table}[t]
    \centering
    \caption{Ablation studies on each component of our method.
    MSL represents Multi-view Scale-invariant Learning; NPL represents Noisy Pseudo box Learning. 
    In MSL, $V_{1}$ and $V_{2}$ are constructed for label- and feature-level consistency, respectively. In NPL, PCV and PLA stand for Positive-proposal Consistency Voting and Prediction-guided Label Assignment, respectively.  
    % \shanshan{It is a bit weird they are not put side by side.}
    }
    % \vspace{-2mm}
    \renewcommand{\arraystretch}{1.1}
    \setlength\tabcolsep{10pt}
    \resizebox{0.75\textwidth}{!}{
    \begin{tabular}{cccc|c|cc}
    \hline
     \multicolumn{2}{c|}{MSL}  & \multicolumn{2}{c|}{NPL}  & \multirow{2}{*}{mAP} & \multirow{2}{*}{AP$_{50}$} & \multirow{2}{*}{AP$_{75}$} \\ \cline{1-4}
    $V_{1}$ & \multicolumn{1}{c|}{$V_{2}$} & PCV & PLA &  &  &   \\ \hline
    &  &  &  & 26.8 & 44.9 & 28.4  \\ \hline
    \checkmark & & & & 33.9{\begin{scriptsize}{(+7.1)}\end{scriptsize}} & 55.2 & 36.0 \\ 
    \checkmark & \checkmark & &  & 34.9{\begin{scriptsize}{(+8.1)}\end{scriptsize}} & 56.3 & 37.1 \\ 
    \checkmark &  & \checkmark & & 34.8{\begin{scriptsize}{(+8.0)}\end{scriptsize}} & 55.1 & 37.4 \\
    \checkmark &  & \checkmark & \checkmark & 35.7{\begin{scriptsize}{(+8.9)}\end{scriptsize}} & 56.4 & 38.4 \\
    \checkmark & \checkmark & \checkmark & &  36.0{\begin{scriptsize}{(+9.2)}\end{scriptsize}} & 56.9 & 38.7 \\
    \checkmark & \checkmark &\checkmark &\checkmark & \textbf{36.3}{\begin{scriptsize}{(+9.5)}\end{scriptsize}} & \textbf{57.2} & \textbf{39.2} \\ \hline
    \end{tabular}}
    \label{tab:components}
\end{table}

\subsection{Ablation Study}

We conduct detailed ablation studies to verify key designs. All ablation studies are conducted on a single data fold from the 10\% labelling ratio.   

\begin{table}[t]
    \centering
    \caption{Analysis of Multi-view Scale-invariant learning, which contains both the label- and feature-level consistency.}
    \vspace{-3mm}
    \begin{subtable}[t]{0.45\textwidth}
    \centering
    \caption{Study on label consistency.}
    \vspace{-1mm}
    \label{tab:label_consistency}
    \resizebox{\linewidth}{!}{
    \begin{tabular}{c|cccc}
    \hline
    method & mAP & AP$_{S}$ & AP$_{M}$ & AP$_{L}$  \\ \hline
    single-scale training & 32.7 & 19.0 & 36.0 & 42.5 \\ 
    label consistency & \textbf{33.9} & \textbf{19.1} & \textbf{37.2} &\textbf{44.4} \\ \hline
    \end{tabular}}
    \end{subtable} \hfill
    \begin{subtable}[t]{0.5\textwidth}
    \centering
    \caption{Study on feature consistency.}
    \vspace{-1mm}
    \label{tab:feature_consistency}
    \resizebox{\linewidth}{!}{
    \begin{tabular}{c|cccc}
    \hline
    method & mAP & AP$_{S}$ & AP$_{M}$ & AP$_{L}$ \\ \hline
    vanilla multi-view training & 33.9 & 20.9 & 37.2 & 43.0 \\
    feature consistency & \textbf{34.9} & \textbf{22.1} & \textbf{38.2} & \textbf{43.6} \\ \hline
    \end{tabular}}
    \end{subtable}
    \label{tab:MSL}
\end{table}

\noindent\textbf{Effect of individual component.} In Tab.~\ref{tab:components}, we show effectiveness of each component step by step. When only using 10\% labeled data as training examples, it obtains 26.8 mAP. Next, we construct the semi-supervised baseline by applying $V_{1}$ on unlabeled data for label-level consistency learning. The baseline does not consider any adverse effects incurred by coarse pseudo boxes and obtains 33.9 mAP. Furthermore, by leveraging additional view $V_{2}$, the feature-level scale-invariant learning is enabled, and an improvement of +1.0 mAP is found. 
On the other hand, to alleviate the issue of coarse pseudo boxes, we introduce PCV to suppress the inaccurate regression signals, improving the baseline from 33.9 to 34.8 mAP. After that, we replace the traditional IoU-based label assignment strategy with the PLA and enjoy another +0.9 mAP gain. 
Finally, when combing MSL and NPL together, it achieves the best performance, 36.3 mAP.

\begin{table*}[t]
    \centering
    \caption{Ablation studies related to Positive-proposal Consistency Voting (PCV) and Prediction-guided Label Assignment (PLA).}
    \vspace{-3mm}
    \begin{subtable}[t]{0.42\textwidth}
    \centering
    \caption{Comparison between our PCV and other regression methods.}
    \vspace{-2mm}
    \label{tab:compare_reg}
    \resizebox{\linewidth}{!}{
    \begin{tabular}{l|ccc}
    \hline     
     method & mAP & AP$_{50}$ & AP$_{75}$   \\ \hline
     abandon reg~\cite{unbiasedTeacher} & 33.9  & 55.2 & 36.0  \\
     reg consistency~\cite{humbleTeacher} & 34.2 & 55.1 & 36.5 \\
     box jittering~\cite{softTeacher} & 34.5 & 54.9 & 36.9 \\ \hline
     PCV (ours) & \textbf{34.8} & \textbf{55.1} & \textbf{37.4}  \\ \hline
    \end{tabular}
    }
    \end{subtable} \hfill
    \begin{subtable}[t]{0.25\textwidth}
    \centering
    \caption{Study on hyper-parameter $\alpha$.}
    \renewcommand{\arraystretch}{1.1}
    \label{tab:study_alpha}
    \resizebox{\linewidth}{!}{
    \begin{tabular}{c|ccc} 
    \hline
    $\alpha$ & mAP & AP$_{50}$ & AP$_{75}$  \\ \hline
        0    & 35.2  &  56.1  &  37.8    \\ 
        0.5  &  \textbf{35.7} & \textbf{56.4}  & \textbf{38.4} \\
        1.0  & 35.4  & 55.7 & 38.4  \\ \hline
    \end{tabular}
    }
    \end{subtable} \hfill
    \begin{subtable}[t]{0.25\textwidth}
    \centering
    \caption{Study on IoU threshold $t$.}
    \label{tab:study_t}
    \renewcommand{\arraystretch}{1.1}
    \resizebox{\linewidth}{!}{
    \begin{tabular}{c|ccc} 
    \hline
    $t$ & mAP & AP$_{50}$ & AP$_{75}$  \\ \hline
    0.3    & 35.7 & 56.2 & \textbf{38.6}   \\ 
    0.4    & \textbf{35.7} & \textbf{56.4}  & 38.4   \\
    0.5    & 35.5 & 56.1 &  38.3   \\ \hline
    \end{tabular}
    }
    \end{subtable} \vfill
\end{table*}

\noindent\textbf{Comparison with other regression methods.} Scores of pseudo boxes can only indicate the confidence of predicted object category, thus they fail to reflect localization quality~\cite{gfl,unbiasedTeacher}. Naive confidence thresholding will introduce some coarse bounding boxes for regression tasks. To alleviate this issue, Unbiased Teacher~\cite{unbiasedTeacher} abandons regression losses on unlabeled data (denoted as ``abandon reg''); 
Humble Teacher~\cite{humbleTeacher} aligns the regression predictions between the teacher and student on selected top-$\mathcal{N}$ proposals (dubbed ``reg consistency''); 
Soft Teacher~\cite{softTeacher} introduces the box jittering to calculate prediction variance on jittered pseudo boxes, which is used to filter out poorly localized pseudo boxes. 
In Tab.~\ref{tab:compare_reg}, we compare our Positive-proposal Consistency Voting (PCV) with these methods.
PCV obtains the best performance, concretely, on AP$_{75}$, PCV surpasses two competitors, reg consistency and box jittering, by 0.9\% and 0.5\%, respectively. 
Although both PCV and box jittering~\cite{softTeacher} rely on prediction variance, there exist great differences. Firstly, PCV produces localization quality by intrinsic proposals, thus it avoids extra network forward on jittered boxes, enjoying higher training efficiency.
Moreover, unlike the box jittering, which meticulously tunes the variance threshold, PCV is free of hyper-parameters.
% Experimental results also show our PCV performs better than box jittering.          

\noindent\textbf{Study on different hyper-parameters of PLA.} We first investigate the performance using different $\alpha$ in PLA, which balances the influence of classification score ($s$) and localization precision ($u$) in the proposal quality. Through a coarse search shown in Tab.~\ref{tab:study_alpha}, we find that combining $s$ and $u$ yields better performance than using them individually.
We then carry out experiments to study the robustness of the IoU threshold $t$, which is used to build the candidate bag. From the Tab~\ref{tab:study_t}, using lower $t$ to construct a bigger candidate bag is preferred.

\noindent\textbf{Analysis of Multi-view Scale-invariant Learning.} \ligang{We propose the MSL to model scale invariance from the aspects of both label- and feature-level consistency.
The studies on them are reported in Tab.~\ref{tab:MSL}. 
At first, we construct a single-scale training baseline without scale variance, where the input images for the teacher and student are kept on the same scale. It obtains 32.7 mAP. Next, we apply the different scale jitter on the teacher and student to implement label-level consistency, which surpasses the single-scale training by 1.2 mAP. 
Based on the label consistency, we further introduce the view $V_{2}$ to perform feature consistency learning. It obtains +1.0\% improvements, reaching 34.9 mAP. Apart from performance gains, the feature consistency can also significantly boost the convergence speed as depicted in Fig.~\ref{fig:convergence}(c). 
To validate the improvements introduced by the $V_{2}$ come from comprehensive scale-invariant learning, instead of vanilla multi-view training, we also add an extra view $V_{2}^{'}$ besides the $V_{1}$, where $V_{2}^{'}$ is downsampled from $V_{1}$ by 2x and performs label consistency as $V_{1}$.
From the Tab.~\ref{tab:feature_consistency}, vanilla multi-view training with only label consistency hardly brings improvements against the single $V_{1}$ (33.9 vs 33.9\%).}

% It is believed that scale-invariant learning plays an important role in consistency training. To validate this, we take single-scale training as the baseline, where no scale-invariant learning is performed and input images for teacher and student are kept on the same scale. Upon this baseline, we first apply label-level consistency (view $V_{1}$), and obtain 1.2\% gains. On medium- and large-scale objects, improvements are significant, specifically, AP$_{M}$/AP$_{L}$ surpasses the baseline by 1.2\%/1.9\%. Furthermore, we combine label- and feature-level consistency together ($V_{1}$\&$V_{2}$), making up our Multi-view Scale-invariant Learning (MSL), and another 1.0\% mAP improvement is found, reaching 34.9\% mAP. Apart from performance improvements, the proposed feature-level consistency can also significantly boost the convergence speed as depicted in Fig.~\ref{fig:convergence}(c). Furthermore, to validate that the improvements introduced by our MSL come from comprehensive scale-invariant learning, instead of naive multi-view training, we introduce another view (named ``$V_{1}$ with low-res''), where input images are downsampled from $V_{1}$ by 2x and label consistency is performed as $V_{1}$. The result is listed in the last row of Tab.~\ref{tab:effects_MSL}.  
\setlength{\belowcaptionskip}{-5mm}
\begin{wraptable}{r}{0.4\linewidth}
\centering
\captionof{table}{Ablation study on Focal Loss.}
\vspace{-3mm}
\label{tab:focal}
\resizebox{0.99\linewidth}{!}{
\begin{tabular}{l|c|cc}
 \hline
 method & mAP & AP$_{50}$ & AP$_{75}$  \\ \hline
 PseCo w/ CE Loss & 35.7 & 55.6 & 38.9 \\
 PseCo w/ Focal Loss & \textbf{36.3} & \textbf{57.2} & \textbf{39.2} \\ \hline
\end{tabular}
}
\end{wraptable}

\noindent\textbf{Effect of Focal Loss.} In Tab.~\ref{tab:focal}, we compare the Cross Entropy (CE) Loss and Focal Loss. Thanks to the Focal Loss, an improvement of +0.6 mAP is achieved against the CE Loss. On the other hand, even with the CE Loss, our PseCo still surpasses the Soft Teacher by a large margin, i.e., 1.7 mAP. 

\section{Conclusion}

In this work, we elaborately analyze two key techniques of semi-supervised object detection (e.g. pseudo labeling and consistency training), and observe these two techniques currently neglect some important properties of object detection. Motivated by this, we propose a new SSOD framework, PseCo, to integrate object detection properties into SSOD. PseCo consists of Noisy Pseudo box Learning (NPL) and Multi-view Scale-invariant Learning (MSL). 
In NPL, prediction-guided label assignment and positive-proposal consistency voting are proposed to perform the robust label assignment and regression task using noisy pseudo boxes, respectively. Based on the common label-level consistency,
MSL additionally designs a novel feature-level scale-invariant learning, which is neglected in prior works. 
%Our MSL combines label- and feature-level consistency, realizing comprehensive scale-invariant learning. 
To validate the effectiveness of our method, extensive experiments are conducted on COCO benchmark. Experimental results validate PseCo surpasses the SOTAs by a large margin both in accuracy and efficiency.

\bibliographystyle{splncs04}
\bibliography{egbib}

\end{document}